\documentclass[conference]{IEEEtran}
\IEEEoverridecommandlockouts
\usepackage{cite}
\usepackage{amsmath,amssymb,amsfonts}
\usepackage{graphicx}
\usepackage{textcomp}
\usepackage{xcolor}
\usepackage{listings}
\usepackage{booktabs}
\usepackage{url}
\usepackage{tikz}
\usetikzlibrary{arrows.meta,positioning}
\usepackage[justification=raggedright,singlelinecheck=false,
            font=small,labelfont=bf,labelsep=period]{caption}

\begin{document}

\title{\LARGE More Accurate or More Efficient? Evaluating Locally Deployed Compact Open-Weight Language Models for Mathematical Reasoning}

\author{\IEEEauthorblockN{Orion Powers}
\IEEEauthorblockA{\textit{College of Engineering and Science} \\
\textit{Florida Institute of Technology}\\
Melbourne, Florida, USA \\
opowers2023@my.fit.edu}
\and
\IEEEauthorblockN{Daniella Seum}
\IEEEauthorblockA{\textit{College of Engineering and Science} \\
\textit{Florida Institute of Technology}\\
Melbourne, Florida, USA \\
dseum2023@my.fit.edu}
\and
\IEEEauthorblockN{Khaled Slhoub}
\IEEEauthorblockA{\textit{College of Engineering and Science} \\
\textit{Florida Institute of Technology}\\
Melbourne, Florida, USA \\
kslhoub@fit.edu}}

\maketitle

\begin{abstract}
Large language models are increasingly deployed on local hardware to address privacy, cost, and accessibility constraints. However, many existing evaluations emphasize model capability and accuracy, while fewer quantify local runtime and energy requirements, characterize failure modes, or apply paired statistical comparisons under controlled conditions. This paper presents a controlled, documented procedure for evaluating locally hosted LLMs on mathematical reasoning. The procedure combines fixed inference settings, hierarchical answer extraction and verification, explicit failure-mode classification, and per-question resource measurement, and reports accuracy using paired significance tests and effect sizes. We demonstrate the procedure in a preliminary study of three compact open-weight models with fewer than five billion parameters, Gemma3:4b (Google), Phi3:3.8b (Microsoft), and Qwen3:4b (Alibaba), across three mathematical reasoning datasets spanning Grade 8 Math, Calculus I, and Advanced Probability and Statistics. All three models were executed through the same local inference server on one workstation, using a shared prompt template, controlled inference settings, and a common matched set of questions per dataset. Results show that no single model dominates. Qwen3:4b achieves the highest accuracy on two datasets, while Gemma3:4b leads on Calculus I. Across every dataset, Gemma3:4b returns roughly three times more correct answers per watt-hour than Qwen3:4b while generating far fewer output tokens per question. Per-question measurements further show that Qwen3:4b requires substantially more generation time and energy, and produces far more output tokens, than Gemma3:4b on every dataset. Phi3:3.8b achieves substantially lower accuracy than the other two models across all three datasets; the low extraction-failure rate indicates that this reflects incorrect answers rather than unparsed output, though we report it with an explicit caveat regarding possible prompt-format effects. These preliminary findings indicate that accuracy alone is an insufficient basis for selecting a local model.
\end{abstract}


\section{Introduction}
Large Language Models (LLMs) have rapidly progressed from research systems to widely used tools in education, research, and software development. As new models continue to be released, users face the challenge of selecting the most appropriate model for their needs. However, model comparisons are often difficult to interpret because providers typically report results using internal benchmarks, while independent evaluations may differ in datasets, metrics, and experimental conditions.

At the same time, locally hosted LLMs have become increasingly important in both industry and academia. Many organizations deploy internal models to maintain data privacy, reduce reliance on external APIs, and control operational costs. Open-weight models can also be deployed through local inference tools such as Ollama \cite{ollama2024} without per-query API charges, making them more accessible to students, researchers, and smaller teams. Despite this shift, many existing benchmarks continue to emphasize broad model capabilities rather than the practical evaluation of smaller locally deployable systems, leaving a gap in evaluation methods for these systems.

A second gap concerns what is measured. When a model runs on someone else's servers, users may not directly observe its hardware-level resource costs. When it runs on user-controlled hardware, the runtime, the number of tokens generated, and the electricity consumed become direct operating considerations, and they can vary substantially between models of similar size. Accuracy alone does not describe whether a model is practical to run locally.

This paper addresses both gaps with a controlled, documented evaluation procedure for locally hosted LLMs, which we demonstrate in a preliminary study of three compact open-weight models in the sub-5B parameter range (fewer than five billion parameters), Gemma3:4b (Google), Phi3:3.8b (Microsoft), and Qwen3:4b (Alibaba), using curated datasets spanning three levels of mathematical complexity: Grade 8 Math, Calculus I, and Advanced Probability and Statistics. Mathematical reasoning was selected because many math problems have objective, verifiable answers, which reduces the need for human or model-based judging. Each model was tested through a controlled procedure that enforces independence between questions, applies consistent prompting, extracts and verifies final answers, and records both correctness and system-level performance metrics.

The contributions of this preliminary study are: (1) a controlled, documented procedure for evaluating locally hosted LLMs on mathematical reasoning that enforces question independence, applies a shared prompt template and fixed inference settings, and records correctness alongside per-question runtime, token, and energy measurements; (2) a multi-stage answer extraction and verification method for mathematical outputs that combines marker-based extraction, pattern and symbolic fallbacks, normalization, and type-aware comparison, reducing reliance on human or model-based judging; (3) an explicit failure-mode classification that separates wrong answers from extraction and execution failures, so that low accuracy can be attributed to reasoning errors rather than parsing problems; (4) a head-to-head comparison of the three models under identical local execution conditions, reported with paired significance testing and effect sizes on accuracy; and (5) the observation that accuracy and efficiency rankings disagree, so that the preferred model depends on the evaluation criterion and intended use case.

We describe this work as preliminary. The study covers three models on one hardware configuration in one task domain, and the underlying evaluation artifacts are not available for release. Section~\ref{sec:threats} states these limitations directly.

\section{Related Work}
Several large-scale benchmark suites characterize LLM capabilities across broad task categories. MMLU measures multitask knowledge across 57 subjects and has become a common reference point for general knowledge and reasoning evaluation \cite{hendrycks2021mmlu}. HELM takes a broader evaluation approach by reporting multiple dimensions of model behavior, including accuracy, calibration, robustness, fairness, bias, toxicity, and efficiency across many scenarios \cite{liang2023helm}. BIG-Bench similarly evaluates language models across a large collection of challenging tasks intended to probe capabilities beyond standard benchmark categories \cite{srivastava2023bigbench}. These suites establish important methodological precedents, particularly in the use of standardized datasets and multi-metric evaluation, but they are primarily designed to evaluate general LLM capability across broad task spaces and do not focus on smaller locally hosted models under identical hardware and inference conditions.

Mathematical reasoning has become a major focus in LLM evaluation because math problems often provide objective answers that can be checked against a known solution. GSM8K is a widely used benchmark of grade-school math word problems requiring multi-step arithmetic reasoning \cite{cobbe2021gsm8k}, and the MATH dataset extends this idea to more advanced competition-level problems \cite{hendrycks2021math}. Continued training on mathematical and technical content, as in Minerva, has been shown to improve quantitative reasoning \cite{lewkowycz2022minerva}, and chain-of-thought prompting can elicit intermediate reasoning steps that improve performance on complex problems, particularly for larger models \cite{wei2022cot}. Mathematical evaluation introduces its own implementation challenges: model outputs may include long reasoning traces, varied notation, equivalent expressions, fractions, decimals, symbolic forms, or final answers embedded in natural language, so direct string comparison is often insufficient for determining correctness.

Efficiency has become an increasingly important consideration in machine learning evaluation. Prior work has documented the financial and environmental costs of large-scale NLP models \cite{strubell2019energy}, and the Green AI proposal argues that efficiency should be treated as a first-class evaluation criterion alongside accuracy rather than as an afterthought \cite{schwartz2020green}. This perspective is especially relevant for locally hosted LLMs, where users directly experience the hardware, power, and runtime costs of inference. More recently, TokenPowerBench measures LLM inference power consumption with phase-aligned attribution of energy to prefill and decode stages \cite{tokenpower2025}, and ConsumerBench evaluates generative AI workloads on end-user devices using application-level latency and system-level utilization and memory-bandwidth metrics \cite{consumerbench2025}. Both measure resource costs in greater depth than we do. Our study is narrower in scope but combines verified final-answer correctness, per-question resource measurements, and paired statistical tests on matched questions in one controlled comparison.

Recent open-weight model families have made local deployment more accessible. Microsoft's Phi series emphasizes compact models designed to achieve strong performance despite small parameter counts, with Phi-3-mini reported as a 3.8B parameter model suitable for resource-constrained deployment \cite{abdin2024phi3}. Google's Gemma family provides lightweight open models built from the research and technology used in Gemini models \cite{gemma2025gemma3}, and Alibaba's Qwen series includes open-weight models across multiple parameter sizes \cite{yang2025qwen3}. These families are commonly evaluated in their own technical reports, but self-reported results are not always directly comparable because they may use different hardware, prompts, datasets, sampling settings, or evaluation procedures. Independent head-to-head comparisons under identical local deployment conditions remain important for understanding how these models perform in practice.

\section{Experimental Methodology}

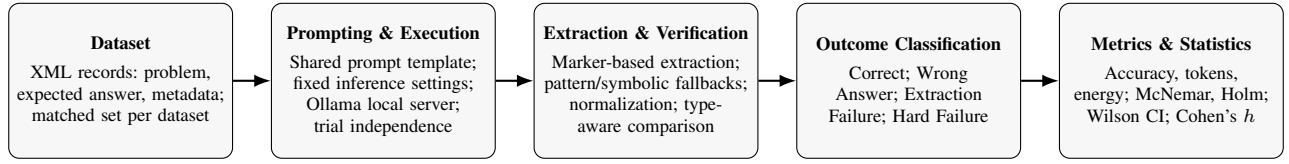
\begin{figure*}[t!]
\centering
\begin{tikzpicture}[
  stage/.style={draw, rounded corners, align=center, text width=2.75cm,
                minimum height=2.05cm, font=\scriptsize, fill=gray!6, inner sep=3pt},
  lbl/.style={font=\scriptsize\bfseries},
  arrow/.style={-{Latex[length=2.2mm]}, thick}
]
\node[stage] (d1) {\textbf{Dataset}\\[3pt] XML records: problem, expected answer, metadata; matched set per dataset};
\node[stage, right=0.5cm of d1] (d2) {\textbf{Prompting \& Execution}\\[3pt] Shared prompt template; fixed inference settings; Ollama local server; trial independence};
\node[stage, right=0.5cm of d2] (d3) {\textbf{Extraction \& Verification}\\[3pt] Marker-based extraction; pattern/symbolic fallbacks; normalization; type-aware comparison};
\node[stage, right=0.5cm of d3] (d4) {\textbf{Outcome Classification}\\[3pt] Correct; Wrong Answer; Extraction Failure; Hard Failure};
\node[stage, right=0.5cm of d4] (d5) {\textbf{Metrics \& Statistics}\\[3pt] Accuracy, tokens, energy; McNemar, Holm; Wilson CI; Cohen's $h$};
\draw[arrow] (d1) -- (d2);
\draw[arrow] (d2) -- (d3);
\draw[arrow] (d3) -- (d4);
\draw[arrow] (d4) -- (d5);
\end{tikzpicture}
\caption{High-level overview of the evaluation procedure. Each stage is applied identically to every model: structured math problems are loaded from the dataset, converted into a standardized prompt and executed on the local inference server, parsed by a multi-stage extraction and verification module, classified into one of four outcome categories, and aggregated into accuracy and per-question resource metrics with paired statistical analysis.}
\label{fig:pipeline}
\end{figure*}

The evaluation procedure is organized as a sequence of controlled stages applied identically to every model: dataset preparation, standardized prompting and local execution, hierarchical answer extraction and verification, and metric collection with statistical analysis. Figure~\ref{fig:pipeline} gives a high-level overview, and each stage is described in the subsections below. The procedure is designed so that the same datasets, prompt template, extraction rules, and metrics can be reapplied to additional models or datasets under the same conditions, which is what makes the resulting comparisons controlled rather than dependent on differing evaluation setups.

\subsection{Datasets}
Three mathematical reasoning datasets were used to evaluate model performance across different educational levels and mathematical domains. The Grade 8 Math dataset contains 644 problems aligned with the categories outlined by the Common Core State Standards: Expressions and Equations, Functions, Geometry, Statistics and Probability, and The Number System. The Calculus I dataset contains 374 problems covering limits, derivatives, integrals, and applications typical of an introductory university calculus course. The Advanced Probability and Statistics dataset contains 884 problems involving probability distributions, statistical inference, and related advanced undergraduate topics.

The datasets were curated from educational sources and organized into structured XML records. In addition to the problem text and expected answer, the records include associated metadata such as category or topic and, where available, directions, solution text, alternate solutions, and difficulty or concept labels. The Grade 8 Math dataset contains one uncategorized problem that is included in the overall accuracy calculation but excluded from the per-category analysis. During dataset preparation, the records were reviewed to confirm that each problem included an expected answer suitable for automated comparison. To support fair pairwise comparisons across models, the analysis uses the matched set of questions evaluated by every model within each dataset, ensuring that each cross-model comparison is based on the same input items. Table~\ref{tab:dataset_summary} summarizes the matched sets.

\begin{table}[htpb!]
\centering
\caption{Matched evaluation set sizes used for cross-model comparison. Grade 8 Math problems are distributed across five Common Core categories; Calculus I and Advanced Probability and Statistics are treated as single-category domains.}
\label{tab:dataset_summary}
\small
\begin{tabular}{lcc}
\toprule
\textbf{Dataset} & \textbf{Matched $N$} & \textbf{Categories} \\
\midrule
Grade 8 Math & 644 & 5 \\
Calculus I & 374 & 1 \\
Advanced Probability and Statistics & 884 & 1 \\
\bottomrule
\end{tabular}
\end{table}

\subsection{Model Selection and Setup}
Three locally deployable sub-5B-parameter LLMs were selected: Gemma3:4b, Phi3:3.8b, and Qwen3:4b. In this naming convention, the numeric suffix refers to the approximate number of parameters in billions. Selecting models within a similar parameter range reduces the influence of large differences in model scale, although the models still differ in architecture, training, release date, and quantization. These models were also selected because they are available for local deployment through Ollama and are among the smaller models offered through that environment, which aligns with the goal of evaluating models that students, researchers, and smaller teams could realistically run without relying on cloud-hosted APIs.

All models were obtained and executed using Ollama, which provides a local inference server and a consistent interface for downloading, loading, and querying open-weight models \cite{ollama2024}. The model artifacts were held fixed throughout the evaluation, and all models were evaluated using the same evaluation pipeline and dataset inputs.

\subsection{Prompt Design and Inference Configuration}
All models received prompts generated from a single shared user-level template, shown in Listing~\ref{lst:prompt}. Each prompt included the mathematical problem text loaded from the dataset and an instruction directing the model to solve the problem and provide a final answer in a consistent format.

\begin{lstlisting}[label=lst:prompt,caption={Shared prompt template used for all models and datasets.}]
Solve the following math problem. Show your work
if helpful, but you MUST end your response with
the final answer on a new line in this exact format:

FINAL_ANSWER: <your answer>

Problem:
<problem_text>
\end{lstlisting}

Evaluation was based on the extracted final answer rather than the reasoning trace. This distinction is important because reasoning text can vary widely between models and may not reliably indicate correctness: a model may provide a correct final answer with limited explanation, or produce detailed reasoning that still leads to an incorrect answer.

The listed inference parameters were held constant across models and evaluation requests. The Ollama inference server was invoked with \texttt{num\_ctx=4096}, \texttt{num\_predict=16384}, \texttt{num\_gpu=999} (full GPU offload), \texttt{num\_thread=8}, \texttt{temperature=0.3}, \texttt{top\_p=0.9}, and \texttt{top\_k=40}. Qwen3:4b additionally received the \texttt{/no\_think} directive to select its non-thinking mode \cite{yang2025qwen3}. This model-specific configuration was used because preliminary evaluations indicated that thinking mode could consume much of the generation budget before the required final-answer marker was produced. The reported Qwen3:4b results therefore characterize its non-thinking configuration; the setting does not make the reasoning processes of the three models equivalent. Section~\ref{sec:discussion} notes the interpretive cost of this choice.

Trial independence was enforced by issuing each query without conversational history, so that prior model responses could not influence later answers. Between models, a 60-second cooldown was applied and the GPU was required to fall below 55$^{\circ}$C before evaluation resumed; before each model run, a fixed warmup prompt (``What is 2+2?'') was issued to ensure model weights were loaded. The warmup request was excluded from the reported results.

\subsection{Answer Extraction and Verification}
A multi-stage extraction and verification procedure was used because LLM outputs are not always formatted consistently. A model may provide a final answer after several lines of reasoning, place the answer inside LaTeX formatting, include units, use equivalent decimal or fractional forms, or express the answer in natural language.

The first stage searches for the expected final answer marker specified in the prompt. If the marker is found, the text following it is extracted as the model's proposed answer. If this primary method fails, fallback rules search for common mathematical answer patterns, LaTeX boxed expressions, numerical values, symbolic expressions, and concluding phrases that may indicate a final answer. The extracted answer is then normalized to handle formatting differences such as whitespace, punctuation, LaTeX syntax, units, fractions, decimals, percentages, coordinates, and intervals, and compared against the expected answer using predefined, deterministic type-aware rules, with a fixed numeric tolerance applied where appropriate. The procedure does not rely on human or model-based judging.

Each response is classified as \emph{Correct} when the normalized answer matches the expected solution, \emph{Wrong Answer} when an answer is successfully extracted but does not match, \emph{Extraction Failure} when the model produces output but no usable final answer can be parsed, and \emph{Hard Failure} when the inference request fails due to timeout, crash, or missing output. This classification distinguishes parsed but incorrect final answers from extraction and execution failures. Extraction failures and hard failures were counted as incorrect when computing accuracy.

\subsection{Metrics and Statistical Analysis}
For each model-question evaluation we record correctness as a binary value, total generation time, prompt and output token counts, and system resource data including GPU utilization, temperature, and power draw. GPU energy consumption is estimated by integrating observed GPU power draw over the inference duration and converting the result into watt-hours. From the per-question records we compute overall accuracy, average output tokens per question, average GPU energy per question, and correct answers per watt-hour. Correct answers per watt-hour is calculated as the total number of correct responses divided by the total estimated GPU energy consumed across the evaluated questions.

Because each model was evaluated on the same matched set of problems, accuracy comparisons were treated as paired comparisons. McNemar's test with Yates' continuity correction was used for pairwise model comparisons \cite{mcnemar1947note}, and accuracy estimates are reported with Wilson 95\% confidence intervals \cite{wilson1927probable}. Holm correction was applied within each dataset across the three model-pair accuracy comparisons to control the family-wise error rate \cite{holm1979simple}, and Cohen's $h$ is reported as a standardized effect-size measure for the differences between marginal accuracies \cite{cohen1988statistical}. Per-question resource metrics (generation time, output tokens, and estimated energy) are summarized descriptively using per-question medians and means. Because the underlying per-question measurements were not retained for release, paired significance tests are not reported for these resource metrics.

\subsection{Experimental Setup}
All evaluations were conducted on a single local workstation running Windows 11 with Ollama as the inference server. The workstation was equipped with a discrete NVIDIA GPU; full GPU layer offload was requested via \texttt{num\_gpu=999}. Peak observed VRAM usage during evaluation was approximately 4.0~GB, peak GPU power draw was approximately 110~W, and system RAM usage peaked at approximately 46~GB. GPU resource metrics were sampled at approximately 2~Hz throughout each query. Before each evaluation, a system validation step verified that GPU utilization was below 5\% and CPU utilization was below 20\% to confirm that no other workloads were competing for resources and that no other Ollama models were resident in memory. The exact hardware identifiers and software versions were not retained; therefore, the absolute runtime and energy measurements should be interpreted as specific to the evaluated workstation and cannot be reproduced exactly.

Models were obtained using the default Ollama tags (\texttt{gemma3:4b}, \texttt{phi3:3.8b}, and \texttt{qwen3:4b}) available at the time of evaluation. Once downloaded, the model artifacts were held fixed throughout the evaluation. The corresponding model digests and quantization identifiers were not retained; therefore, subsequent versions associated with these tags may differ from those evaluated in this study. Evaluations were conducted between April~6 and April~17, 2026.

\section{Preliminary Results}

\subsection{Overall Accuracy}
Table~\ref{tab:main} reports overall accuracy for each model and dataset combination, along with Wilson 95\% confidence intervals. Qwen3:4b achieved the highest accuracy on Advanced Probability and Statistics (0.974) and Grade 8 Math (0.783), while Gemma3:4b achieved the highest accuracy on Calculus I (0.864). The results also show a clear performance gap between Phi3:3.8b and the other two models across all datasets, with Phi3:3.8b scoring 0.104 on Advanced Probability and Statistics, 0.289 on Calculus I, and 0.025 on Grade 8 Math. Model rankings therefore differ across mathematical domains.

\begin{table*}[htpb!]
\centering
\caption{Accuracy and efficiency on the matched evaluation set. $N$ denotes the number of questions in the matched set. Gemma3:4b leads on correct answers per watt-hour for every dataset despite being outscored on raw accuracy for two of three. Qwen3:4b generates roughly three to four times more output tokens per question than Gemma3:4b.}
\label{tab:main}
\begin{tabular}{llccccc}
\toprule
\textbf{Model} & \textbf{Dataset} & \textbf{$N$} & \textbf{Accuracy} & \textbf{95\% CI} & \textbf{Correct/Wh} & \textbf{Tokens/Q} \\
\midrule
Gemma3:4b & Adv.\ Prob.\ \& Stat. & 884 & 0.947 & [0.930, 0.960] & 6.53 & 440 \\
Phi3:3.8b & Adv.\ Prob.\ \& Stat. & 884 & 0.104 & [0.086, 0.126] & 0.89 & 389 \\
Qwen3:4b & Adv.\ Prob.\ \& Stat. & 884 & 0.974 & [0.961, 0.983] & 1.96 & 1{,}507 \\
\midrule
Gemma3:4b & Calculus I & 374 & 0.864 & [0.825, 0.895] & 4.45 & 502 \\
Phi3:3.8b & Calculus I & 374 & 0.289 & [0.245, 0.337] & 0.96 & 1{,}189 \\
Qwen3:4b & Calculus I & 374 & 0.794 & [0.750, 0.832] & 1.62 & 1{,}485 \\
\midrule
Gemma3:4b & Grade 8 Math & 644 & 0.675 & [0.638, 0.710] & 4.84 & 361 \\
Phi3:3.8b & Grade 8 Math & 644 & 0.025 & [0.015, 0.040] & 0.20 & 401 \\
Qwen3:4b & Grade 8 Math & 644 & 0.783 & [0.749, 0.813] & 1.42 & 1{,}518 \\
\bottomrule
\end{tabular}
\end{table*}

To determine whether the observed accuracy differences were statistically reliable, McNemar's test with continuity correction was applied to every pair of models on each dataset, with $p$-values adjusted using Holm correction across the three model-pair comparisons for each dataset. Table~\ref{tab:pairwise_accuracy} reports the results. All nine pairwise comparisons are statistically significant at $\alpha = 0.05$. The comparisons involving Phi3:3.8b show large accuracy differences and large Cohen's $h$ effect sizes, between 1.06 and 2.16 in magnitude. The comparisons between Gemma3:4b and Qwen3:4b remain significant on each dataset but are far smaller in practical terms, with $h = -0.14$ on Advanced Probability and Statistics, $+0.19$ on Calculus I, and $-0.24$ on Grade 8 Math. For these three comparisons, statistical significance should not be interpreted as evidence of a large practical difference.

\begin{table*}[htpb!]
\centering
\caption{Pairwise accuracy comparisons via McNemar's test with Holm correction on the matched evaluation set. All nine comparisons are statistically significant at $\alpha = 0.05$. A positive accuracy difference (A$-$B) indicates Model A is more accurate; Cohen's $h$ reports effect-size magnitude.}
\label{tab:pairwise_accuracy}
\begin{tabular}{llcccc}
\toprule
\textbf{Dataset} & \textbf{Pair (A vs.\ B)} & \textbf{Acc.\ Diff.} & \textbf{$p$ (Holm)} & \textbf{$h$} & \textbf{Mag.} \\
\midrule
Adv.\ P\&S & Gemma3 vs.\ Phi3 & $+0.843$ & $< 10^{-162}$ & $+2.02$ & large \\
Adv.\ P\&S & Gemma3 vs.\ Qwen3 & $-0.027$ & 0.002 & $-0.14$ & negl. \\
Adv.\ P\&S & Phi3 vs.\ Qwen3 & $-0.870$ & $< 10^{-160}$ & $-2.16$ & large \\
\midrule
Calc.\ I & Gemma3 vs.\ Phi3 & $+0.575$ & $< 10^{-44}$ & $+1.25$ & large \\
Calc.\ I & Gemma3 vs.\ Qwen3 & $+0.070$ & $3.1 \times 10^{-4}$ & $+0.19$ & negl. \\
Calc.\ I & Phi3 vs.\ Qwen3 & $-0.505$ & $< 10^{-40}$ & $-1.06$ & large \\
\midrule
Grade 8 & Gemma3 vs.\ Phi3 & $+0.651$ & $< 10^{-90}$ & $+1.61$ & large \\
Grade 8 & Gemma3 vs.\ Qwen3 & $-0.107$ & $< 10^{-13}$ & $-0.24$ & small \\
Grade 8 & Phi3 vs.\ Qwen3 & $-0.758$ & $< 10^{-105}$ & $-1.85$ & large \\
\bottomrule
\end{tabular}
\end{table*}

Within Grade 8 Math, Qwen3:4b achieved the highest accuracy in all five Common Core categories, and Gemma3:4b was second in each. Table~\ref{tab:grade8_category} reports the per-category breakdown. The five categories cover 643 of the 644 problems; one problem is uncategorized in the source dataset and is included in the overall accuracy but not in this table. Phi3:3.8b performed substantially lower across all five categories, reaching its highest category-level accuracy, 0.091, on The Number System. Both Gemma3:4b and Qwen3:4b achieved their lowest category-level accuracies on Expressions and Equations (0.505 and 0.591, respectively), indicating a shared weakness that overall accuracy obscures. The two models were closest on Statistics and Probability and differed most on Geometry, followed by Functions.

\begin{table*}[htbp!]
\centering
\caption{Per-category accuracy on Grade 8 Math. Qwen3:4b leads every category; the two stronger models are closest on Statistics and Probability and both struggle on Expressions and Equations.}
\label{tab:grade8_category}
\begin{tabular}{lcccc}
\toprule
\textbf{Category} & \textbf{$N$} & \textbf{Gemma3:4b} & \textbf{Phi3:3.8b} & \textbf{Qwen3:4b} \\
\midrule
Expressions and Equations & 186 & 0.505 & 0.005 & 0.591 \\
Functions & 88 & 0.864 & 0.000 & 0.977 \\
Geometry & 145 & 0.690 & 0.000 & 0.876 \\
Statistics and Probability & 60 & 0.933 & 0.000 & 0.950 \\
The Number System & 164 & 0.659 & 0.091 & 0.750 \\
\bottomrule
\end{tabular}
\end{table*}

\subsection{Energy and Efficiency}
Accuracy alone does not fully describe the performance of a locally deployed model. For local execution, energy usage, token generation, and runtime cost are also important because they affect the practicality of running the model on available hardware.

Gemma3:4b achieved the highest correct answers per watt-hour on all three datasets, reaching 6.53 on Advanced Probability and Statistics, 4.45 on Calculus I, and 4.84 on Grade 8 Math, against 1.96, 1.62, and 1.42 for Qwen3:4b. A major factor associated with this difference is output length. Qwen3:4b averaged 1,507, 1,485, and 1,518 tokens per question on the three datasets, while Gemma3:4b averaged 440, 502, and 361. This difference is reflected in energy per question, where Qwen3:4b consumed a mean of 0.496~Wh, 0.490~Wh, and 0.553~Wh per question, while Gemma3:4b consumed 0.145~Wh, 0.194~Wh, and 0.140~Wh.

Table~\ref{tab:resource_medians} reports per-question medians for generation time, output tokens, and energy on each dataset. The same pattern holds across all three datasets: Qwen3:4b is by a wide margin the most resource-intensive of the three models, with median per-question energy roughly three times that of Gemma3:4b and about three times as many median output tokens, mirroring its much longer generation times. Gemma3:4b and Phi3:3.8b are far closer to each other, and their ordering depends on the dataset. On Advanced Probability and Statistics and Grade 8 Math, Gemma3:4b is slower and consumes more energy per question than Phi3:3.8b. On Calculus I, however, the two models are comparable in generation time, and Phi3:3.8b in fact produces more median output tokens and consumes slightly more energy per question than Gemma3:4b, consistent with the higher mean output-token count for Phi3:3.8b on Calculus I in Table~\ref{tab:main}. Phi3:3.8b is therefore the least costly model per question on two of the three datasets, but any efficiency advantage is offset by its much lower accuracy. Because the underlying per-question measurements were not retained, these resource results are reported descriptively rather than with paired significance tests. The accuracy comparisons reported above, which rest on the retained correctness data, are unaffected.

\begin{table}[htpb!]
\centering
\caption{Per-question resource medians on the matched evaluation set: generation time, output tokens, and estimated GPU energy for each model. Qwen3:4b is the most resource-intensive model across all three datasets. Phi3:3.8b has the lowest resource medians on Advanced Probability and Statistics and Grade 8 Math, whereas Gemma3:4b has the lowest medians on Calculus I. Corresponding per-question mean output-token values appear in Table~\ref{tab:main}, and per-question mean energy values are reported in the text.}

\label{tab:resource_medians}
\small
\setlength{\tabcolsep}{5pt}
\begin{tabular}{lccc}
\toprule
\textbf{Metric} & \textbf{Gemma3:4b} & \textbf{Phi3:3.8b} & \textbf{Qwen3:4b} \\
\midrule
\multicolumn{4}{l}{\textit{Advanced Probability and Statistics}} \\
Gen.\ Time (s) & 6.79 & 4.36 & 18.25 \\
Output Tokens  & 431  & 240  & 1{,}261 \\
Energy/Q (Wh)  & 0.142 & 0.075 & 0.423 \\
\midrule
\multicolumn{4}{l}{\textit{Calculus I}} \\
Gen.\ Time (s) & 9.26 & 9.33 & 19.38 \\
Output Tokens  & 421  & 856  & 1{,}277 \\
Energy/Q (Wh)  & 0.159 & 0.210 & 0.416 \\
\midrule
\multicolumn{4}{l}{\textit{Grade 8 Math}} \\
Gen.\ Time (s) & 6.89 & 4.17 & 16.30 \\
Output Tokens  & 322  & 217  & 1{,}131 \\
Energy/Q (Wh)  & 0.127 & 0.073 & 0.416 \\
\bottomrule
\end{tabular}
\end{table}

\subsection{Failure Modes}
Table~\ref{tab:failure_modes} reports the failure-mode breakdown. Extraction failures were negligible. Gemma3:4b and Qwen3:4b had 0.0\% extraction failure rates across all datasets, and Phi3:3.8b failed extraction on only 0.16\% of Grade 8 Math questions and 0.0\% elsewhere. No hard failures occurred. Because extraction failures were rare, most incorrect responses were classified as wrong answers rather than formatting or parsing failures, which indicates that the reported accuracy differences were not driven by unparsed responses. The distinction matters most for Phi3:3.8b: its wrong-answer rate reaches 97.4\% on Grade 8 Math, showing that its low accuracy came from incorrect answers the pipeline parsed successfully, not from an inability to extract its output.

\begin{table*}[htbp!]
\centering
\caption{Failure-mode classification on the matched evaluation set. Wrong Answer Rate is the fraction of all evaluated responses for which an answer was parsed but determined to be incorrect. Extraction Failure Rate is the fraction of all evaluated responses from which no answer could be parsed. Extraction failures are negligible across all cells.}
\label{tab:failure_modes}
\begin{tabular}{llccc}
\toprule
\textbf{Model} & \textbf{Dataset} & \textbf{Wrong Ans.} & \textbf{Extr.\ Fail.} & \textbf{Hard Fail.} \\
\midrule
Gemma3:4b & Adv.\ P\&S & 5.3\% & 0.0\% & 0.0\% \\
Phi3:3.8b & Adv.\ P\&S & 89.6\% & 0.0\% & 0.0\% \\
Qwen3:4b & Adv.\ P\&S & 2.6\% & 0.0\% & 0.0\% \\
\midrule
Gemma3:4b & Calculus I & 13.6\% & 0.0\% & 0.0\% \\
Phi3:3.8b & Calculus I & 71.1\% & 0.0\% & 0.0\% \\
Qwen3:4b & Calculus I & 20.6\% & 0.0\% & 0.0\% \\
\midrule
Gemma3:4b & Grade 8 Math & 32.5\% & 0.0\% & 0.0\% \\
Phi3:3.8b & Grade 8 Math & 97.4\% & 0.16\% & 0.0\% \\
Qwen3:4b & Grade 8 Math & 21.7\% & 0.0\% & 0.0\% \\
\bottomrule
\end{tabular}
\end{table*}

\section{Discussion}
\label{sec:discussion}

\subsection{Accuracy and Efficiency Diverge}
The results show that accuracy and energy efficiency do not align among the three models evaluated. The best-performing model depends on the metric being prioritized. This distinction matters for local deployment because locally hosted models are executed directly on user hardware. In cloud-based systems, users may only see cost in terms of API pricing or response time. In local systems, the user is also affected by hardware load, power usage, runtime, and the number of queries being executed.

When raw accuracy is prioritized, Qwen3:4b achieved the highest accuracy on Advanced Probability and Statistics and Grade 8 Math, whereas Gemma3:4b achieved the highest accuracy on Calculus I (Table~\ref{tab:main}). For workloads involving many queries, cumulative energy and runtime differences may make Gemma3:4b more attractive, although the accuracy consequences remain dataset-dependent. Neither model is uniformly preferable across evaluation criteria and mathematical domains. The results therefore do not support selecting a model based only on accuracy or parameter count: all three models were in a similar parameter range, but their accuracy, energy efficiency, output length, and failure patterns differed substantially.

\subsection{Interpreting the Phi3:3.8b Result}
Phi3:3.8b performed substantially lower than the other two models on every dataset, scoring 0.025 on Grade 8 Math. This result is substantially lower than the performance reported for Phi-3-mini on GSM8K. However, the datasets, prompting strategies, and evaluation conditions differ, preventing a direct comparison \cite{abdin2024phi3}. We report this result but do not claim it reflects the model's mathematical capability.

The low extraction failure rate indicates that the verification procedure extracted a candidate answer from nearly all of the model's outputs, but parseable output is not evidence of a well-posed prompt. A mismatch between our shared prompt template and the chat format Phi3:3.8b expects could produce fluent, extractable, and incorrect responses, and we did not perform the transcript-level inspection needed to exclude that explanation. Readers should treat these figures as characterizing this specific model under this specific evaluation procedure rather than as a general capability estimate. Determining which explanation holds is the first priority of future work, and the answer bears directly on how fairly any fixed-prompt evaluation treats heterogeneous models.

One additional contextual factor is the relative age of the models. Phi3:3.8b was released in April 2024 \cite{abdin2024phi3}, while Gemma3:4b and Qwen3:4b were released in March and April 2025, respectively \cite{gemma2025gemma3,yang2025qwen3}. Release age alone should not be interpreted as an explanation for the observed performance difference. Since LLM development progresses rapidly, results should be interpreted as a comparison of the specific Ollama model instances and configurations evaluated rather than a general ranking of the Phi, Gemma, or Qwen model families.

\subsection{The Cost of Disabling Deliberation}
Qwen3:4b was evaluated with thinking mode disabled \cite{yang2025qwen3}. This choice prevented internal deliberation from consuming the available token budget before a visible final answer was produced and provided a response configuration more comparable to those of Gemma3:4b and Phi3:3.8b. However, disabling thinking may have affected Qwen3:4b's accuracy and reduced its resource consumption relative to its thinking-enabled configuration because deliberation would require additional token generation. The accuracy and efficiency tradeoff reported here therefore describes Qwen3:4b in its non-deliberative configuration, and an explicit comparison across both configurations remains open.

\section{Threats to Validity and Limitations}
\label{sec:threats}

\textbf{Artifact availability.} The evaluation code and processed datasets are not available for public release. However, the paper documents the dataset sources and matched-set sizes, prompt template, inference parameters, evaluation procedure, metrics, and statistical methods to support an independent replication of the study design. Because the exact hardware and software identifiers and Ollama model digests were not retained, a replication may not reproduce the reported numerical values exactly.

\textbf{Dataset validity and contamination.} The datasets were curated from educational sources and reviewed to ensure each record included a valid expected answer, but some residual label noise may remain, including ambiguous wording, alternate valid answer forms, and expected answers that do not capture every mathematically equivalent response. Separately, Common Core-aligned problems and standard calculus exercises are heavily represented online and may appear in the models' training data, so absolute accuracy values, particularly the 0.974 on Advanced Probability and Statistics, may be inflated. Presenting identical matched datasets to all models controls differences in question selection but does not eliminate contamination or the possibility that training-data exposure differed among models. Therefore, contamination may also affect the relative comparisons.

\textbf{Construct validity.} Correctness is measured by comparing the extracted final answer against the expected answer, which does not evaluate the quality of the model's reasoning process. Automated extraction, normalization, and matching may also misclassify some responses, particularly when mathematically equivalent answers use unexpected formats. Energy consumption is estimated by integrating GPU power draw, which captures only the GPU component of local inference energy and excludes CPU, host memory, storage, and cooling. Power was sampled at approximately 2~Hz, so short queries yield few samples and per-question estimates for fast models carry proportionally more error. Applying the same measurement procedure to all models improves comparability but does not eliminate this measurement error.

\textbf{Internal validity.} All evaluations were performed on a single local workstation. Although background system load was minimized and the same software environment was used throughout, hardware-specific factors including GPU driver behavior, thermal conditions, and operating system scheduling may still influence timing and energy measurements. Absolute runtime and energy values should be interpreted as characteristic of the specific workstation used. Results also reflect one prompt template and one sampling configuration; a different prompt or configuration could affect both absolute accuracy and relative model rankings, particularly if the shared prompt interacts differently with each model. The available documentation does not specify the number of repeated runs or how measurements across runs were aggregated, limiting assessment of run-to-run variability.

\textbf{External validity.} The results are most directly applicable to locally hosted, sub-5B-parameter open-weight models evaluated on mathematical reasoning tasks through Ollama. They should not be assumed to generalize to larger models, proprietary cloud-hosted systems, non-mathematical tasks, or different inference frameworks. The models were obtained using default Ollama tags, but their exact digests and quantization identifiers were not retained. Because registry tags may change over time, the results apply only to the model artifacts available during the evaluation period and may differ across later versions or other quantization levels.

\section{Conclusion and Future Work}
This paper reported a preliminary comparison of three sub-5B-parameter open-weight models on three mathematical reasoning datasets under controlled local execution conditions, recording accuracy alongside runtime, output tokens, and estimated energy per question. The main finding is that local LLM performance cannot be fully described by accuracy alone. Qwen3:4b achieved the highest accuracy on Advanced Probability and Statistics and Grade 8 Math, while Gemma3:4b achieved the highest accuracy on Calculus I and returned roughly three times more correct answers per watt-hour than Qwen3:4b on every dataset. Paired tests identified statistically significant accuracy differences between all three models. For the comparison between Gemma3:4b and Qwen3:4b specifically, the accuracy difference is negligible to small in effect-size terms, whereas the per-question resource gap is large: Qwen3:4b generates roughly two-and-a-half to four times more output tokens and consumes roughly three to four times as much GPU energy per question as Gemma3:4b, depending on the dataset. These findings show that similarly sized local models can differ substantially in accuracy, efficiency, and domain-specific performance

Future work has four priorities. First, transcript-level inspection of Phi3:3.8b to investigate whether its low accuracy reflects model capability or prompt-format mismatch. Second, direct evaluation of Qwen3:4b with thinking mode enabled and disabled to quantify its effect on accuracy and resource consumption. Third, development of a publicly released and reproducible implementation that preserves versioned model identifiers, complete hardware and software specifications, per-question results, and sufficient repeated runs to characterize run-to-run variance. Fourth, extension beyond mathematical reasoning to code, scientific, and natural language tasks, together with stronger symbolic-equivalence checking and evaluation across multiple hardware configurations to characterize the contribution of hardware variability to runtime and energy measurements.


\end{document}